\documentclass[11pt]{article}
\usepackage{amsfonts}

\usepackage[final]{acl}

\usepackage{times}
\usepackage{latexsym}

\usepackage[T1]{fontenc}

\usepackage[utf8]{inputenc}

\usepackage{microtype}

\usepackage{inconsolata}

\usepackage{graphicx}
\usepackage{multirow}
\usepackage{amsmath}
\usepackage{booktabs}
\usepackage{tcolorbox}
\usepackage[table]{xcolor} 
\usepackage{}
\newtcolorbox{prompt}[1]{colback=gray!5,colframe=gray!35!black,fonttitle=\bfseries, title={#1}}
\newcommand{\ours}[1]{\textsc{OMGuard}}
\newcommand{\ds}[1]{\textsc{MM-Misleading}}

\usepackage{xcolor, soul,todonotes} 
	\definecolor{kmycolor}{rgb}{0.858, 0.188, 0.478}

\title{What's Left Unsaid? Detecting and Correcting \\ Misleading Omissions in Multimodal News Previews}

\author{Fanxiao Li\textsuperscript{1}, Jiaying Wu\textsuperscript{2}, 
Tingchao Fu\textsuperscript{1}, 
Dayang Li\textsuperscript{1}, 
Herun Wan\textsuperscript{3}, \\
\textbf{Wei Zhou\textsuperscript{4} \thanks{Corresponding Author}, 
Min-Yen Kan\textsuperscript{2}}\\
  \textsuperscript{1} School of Information Science and Engineering, Yunnan University,\\
   \textsuperscript{2} National University of Singapore,
   \textsuperscript{3} Xi’an Jiaotong University, \\
   \textsuperscript{4} School of Engineering, Yunnan University \\
    \texttt{lifanxiao@stu.ynu.edu.cn, jiayingwu@u.nus.edu, zwei@ynu.edu.cn}
  }

\begin{document}
\maketitle

\begin{abstract}
Even when factually correct, social-media news previews (image--headline pairs) can induce \textbf{interpretation drift}: by selectively \textit{omitting} crucial context, they lead readers to form judgments that diverge from what the full article supports. This covert harm is subtler than explicit misinformation, yet remains underexplored. To address this gap, we develop a multi-stage pipeline that simulates preview-based and context-based understanding, enabling construction of the \ds{} benchmark. Using \ds{}, we systematically evaluate open-source LVLMs and uncover pronounced blind spots in \textit{omission-based misleadingness detection}. We further propose \ours{}, which combines \textbf{(1) Interpretation-Aware Fine-Tuning} for misleadingness detection and \textbf{(2) Rationale-Guided Misleading Content Correction}, where explicit rationales guide headline rewriting to reduce misleading impressions. Experiments show that \textsc{OMGuard} lifts an 8B model’s detection accuracy to the level of a 235B LVLM while delivering markedly stronger end-to-end correction. Further analysis shows that misleadingness usually arises from local narrative shifts, such as missing background, instead of global frame changes, and identifies image-driven cases where text-only correction fails, underscoring the need for visual interventions.\footnote{Data and code are available at \url{https://github.com/fanxiao15/OMGuard}.}

\end{abstract}

\section{Introduction}

\begin{figure}[t!]
\begin{center}
    \includegraphics[width=\linewidth]  {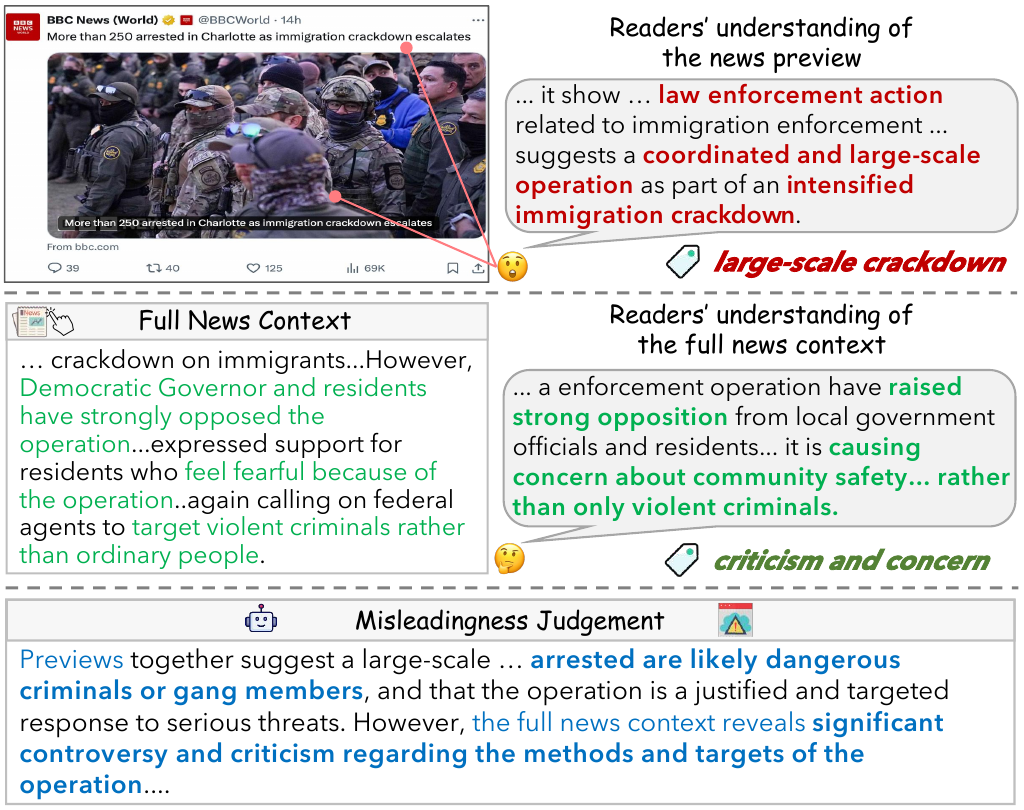}
    \caption{\textbf{Illustration of misleading omissions in multimodal news previews}. Social media users often encounter only a \textit{news preview} (image--headline pair), while the \textit{full context} remains unavailable unless they click through. When key information is omitted or selectively presented, the preview can induce interpretations that diverge from those supported by the full article.}    
    \label{fig:intro}
    \vspace{-4mm} 
\end{center}
\end{figure}







The governance of online misinformation is evolving beyond the detection of explicit fabrications. While existing efforts primarily target factually incorrect content~\cite{qi2024sniffer, wan2025truth}, a more insidious risk arises from \textbf{omission-based deception}~\cite{van2014communicating,allen2024quantifying}, where creators subtly reshape impressions by omitting essential background context. This risk is intensified on social media, where users often rely on \textbf{news previews} (a brief headline and image), and more than 75\% of links are shared without a click~\cite{sundar2025sharing}. When the interpretation formed from the preview significantly diverges from the narrative supported by the full article, it creates what we define as \textbf{misleading omissions}. As illustrated in Figure~\ref{fig:intro}, a news preview may frame the event as a coordinated crackdown on illegal actors, even though the full article highlights public concerns about overly aggressive operations affecting ordinary immigrants.

Despite growing interest in misinformation governance, existing methods offer limited support for this challenge. Many approaches emphasize fact verification and treat omission primarily as an auxiliary cue~\cite{tracer, OmiGraph}. Other lines of work study creator intent~\cite{wu2025seeing, NINT, Inside}, framing strategies~\cite{arora2025multi_framing,moernaut2020visual,lucking2012framing,ali2022survey}, or clickbait engagement~\cite{wang2025multi, yu2024multimodalclick, hagen2022clickbait}. These methods do not examine whether preview-level omissions create a \textit{systematic mismatch} relative to the full context, and they provide no mechanisms to proactively mitigate this effect through preview correction.

To bridge this gap, we develop \ds{}, a benchmark of 6,000 multimodal news previews derived from VisualNews~\cite{visualnews}. Constructed through a cognitive simulation pipeline, \ds{} captures the interpretive divergence between preview and full context, providing omission labels, supporting rationales, and high-quality corrections. Cross-model agreement filtering ensures annotation reliability, and human evaluation confirms substantial alignment with expert judgment. This establishes \ds{} as a rigorous testbed for both detection and mitigation. However, our evaluation of representative open-source LVLMs on \ds{} uncovers a fundamental weakness: the models often fail to compare the preview against the full context, leading to systematic failures to detect the subtle omission-induced shifts that define misleadingness.

Motivated by these findings, we propose \ours{}, a framework for detecting and mitigating misleading omissions. At its core is \textit{interpretation-aware fine-tuning}, which contrasts preview-based and context-based interpretations to strengthen small models’ ability to detect and explain omission-based misleadingness. Beyond detection, \ours{} uses its generated rationales to support \textbf{active mitigation}: the model performs minimally invasive edits to the headline to reduce misleading interpretations relative to the full context. This allows \ours{} to function as both \textbf{(1)} a platform-side screening tool for warning generation and \textbf{(2)} a publisher-side newsroom copilot for pre-publication self-checks.

Extensive experiments on \ds{} validate the effectiveness of our approach. On both omission detection and mitigation, \ours{} based on an 8B model matches or surpasses the performance of a strong 235B LVLM. We identify a \textit{perception–correction gap} in omission correction, where end-to-end correction success critically depends on upstream detection and the quality of generated rationales, and we demonstrate that explicit interpretation guidance produces significant zero-shot transfer gains. Our analysis further shows that omission-based misleadingness typically manifests as local narrative shifts, such as missing background information, rather than global reframing. For image-driven cases where text-only correction is insufficient, we also show that \ours{} supports visual prototyping, a necessary strategy for fully multimodal mitigation.

\section{Related Work}
\paragraph{Evidence-Based Fact Checking.}
Most existing work~\cite{qi2024sniffer, li-etal-2025-cmie, wu2025e2lvlm, wan2025difar} focuses on fact-checking, leveraging retrieved external evidence to assess the factual correctness of image–text pairs. Recent studies~\cite{tracer, OmiGraph} have also examined omissions, but they mainly retrieve missing information to infer implied intent and use it as an auxiliary signal for fact verification. In contrast, we study a different yet underexplored form of deception: content that remains factually correct in the full context but becomes misleading due to critical omissions at the preview level.

\paragraph{News Preview Understanding Beyond Veracity.}
Recent studies have also examined how the \emph{presentation} of news shapes audience understanding and dissemination outcomes. DeceptionDecoded~\cite{wu2025seeing} analyzes misleading presentations constructed by creators under different intents, while NINT and InSide~\cite{NINT,Inside} leverage intent as an auxiliary signal for misinformation detection. Journalism research on framing~\cite{de2005news,arora2025multi_framing,moernaut2020visual} focuses on identifying and attributing frame categories, and clickbait studies~\cite{wang2025multi,yu2024multimodalclick,hagen2022clickbait} investigate how previews influence clicks and engagement. However, none of these lines of work explicitly characterizes the reader understanding gap induced between the preview and the full context. 

A fuller discussion of related work is provided in Appendix~\ref{app:related_work}.

\section{Problem Definition}
\label{sec:problem_definition}

\subsection{Misleading Omission Detection}
\label{sec:misleading_omission_detection}
Let $P_{\text{news}} = (T, I)$ denote a multimodal news preview, where $T$ is a one-sentence headline and $I$ is the accompanying image. The preview links to a full news article $C_{\text{news}}$. The goal is to detect misleading omissions in $P_{\text{news}}$, where selective presentation creates an impression that deviates from reading $C_{\text{news}}$. 

Motivated by prior work on news comprehension~\cite{entman1993framing, rich2016continued, petty1986elaboration}, we simulate the reader’s understanding under preview-only and full-context exposure by constructing two intermediate interpretations:
\begin{equation}
U_p = h_{\theta}(P_{\text{news}}), \qquad
U_c = h_{\theta}(C_{\text{news}}).
\end{equation}
These serve as reasoning states that capture preview-based and context-based impressions.

The model then predicts a binary label $y \in {0,1}$ and a rationale $r$:
\begin{equation}
y, r = h_{\theta}\big(P_{\text{news}}, C_{\text{news}}, U_p, U_c \big).
\end{equation}

A preview is labeled as misleading ($y=1$) if discrepancies between $U_p$ and $U_c$ result in a substantial shift in the perceived nature or implications of the event; otherwise it is labeled as non-misleading ($y=0$). The rationale $r$ explains the source and effect of the omission.


\subsection{Misleading Omission Correction}
\label{sec:misleading_omission_correction}

Given a misleading instance and its rationale $((T, I), C_{\text{news}}, r)$, the goal is to correct the preview by eliminating misleading omissions. 
Since news images serve as documentary evidence and editing them may introduce fabrication or reduce authenticity, we fix the image $I$ as an anchor and focus on \textit{headline revision}: generating a revised headline $\hat{T}$ so that the preview $(\hat{T}, I)$ accurately reflects the full context $C_{\text{news}}$.


To ensure practical usability, we impose a strict length budget, allowing at most three additional words. Under this constraint, we introduce two correction protocols (see Figure~\ref{prompt:headline_correction}) that emphasize \emph{style preservation} versus \emph{factual restoration}:
\textbf{(1) Minimal-Edit Revision}: prioritizes \textit{editorial continuity} by preserving the original style, tone, and syntactic structure as much as possible, simulating a newsroom-copilot setting where the preview must remain close to its published form;
\textbf{(2) Free-Form Revision}: prioritizes informational fidelity by relaxing stylistic constraints and focusing solely on eliminating misleading impressions, approximating an upper bound of correction when optimized for truthfulness. The correction process is formalized as:
\begin{equation}
\label{eq:mitigation}
\hat{T} = h_{\theta}\big((T, I), C_{\text{news}}, r, \mathcal{C}\big),
\end{equation}
where $\mathcal{C}$ denotes the chosen correction protocol.

\section{The \ds{} Benchmark}
\label{sec:misleading_benchmark}

\begin{figure*}[t!]
\begin{center}
    \includegraphics[width=\linewidth]  {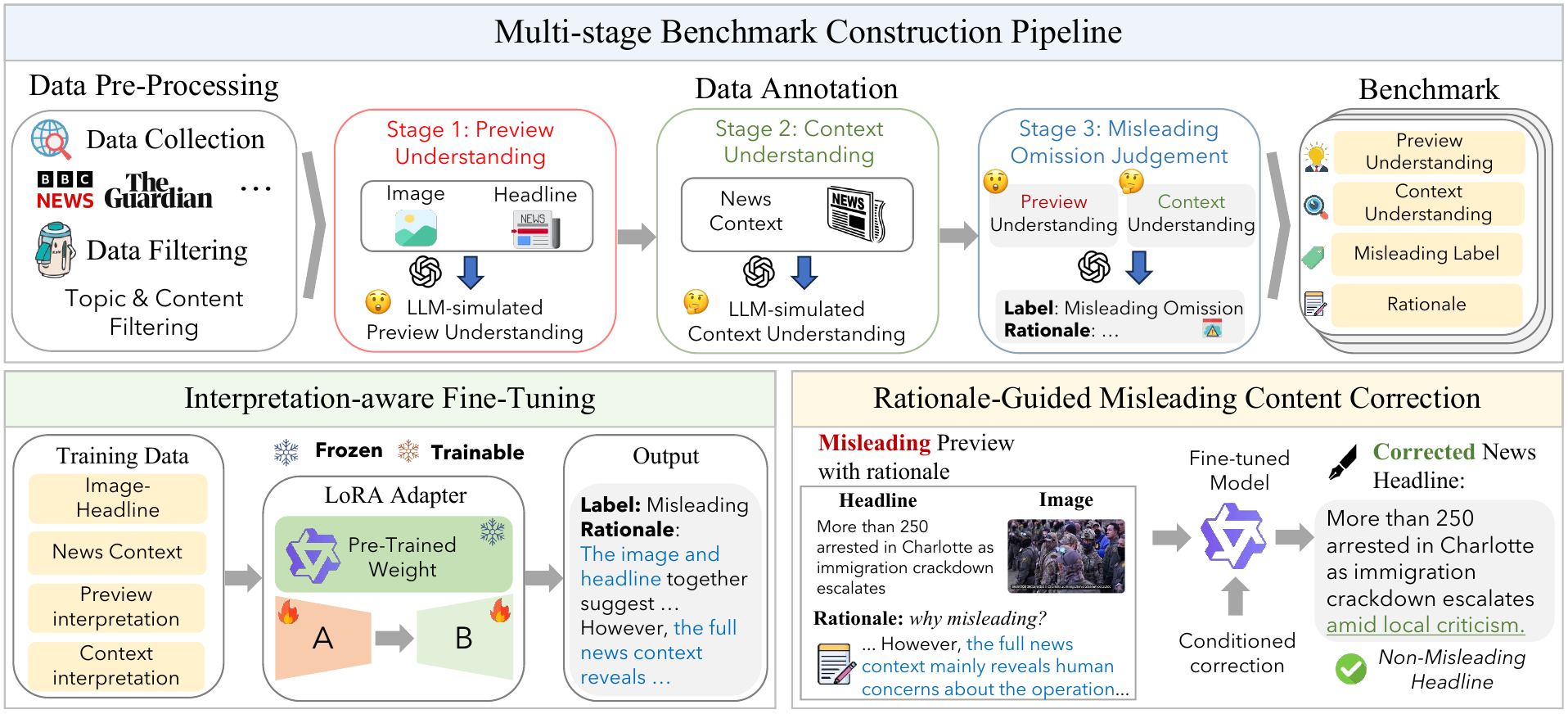}
    \caption{\textbf{Overview of \ours{}.} The upper section illustrates the multi-stage annotation pipeline described in \S~\ref{sec:data_annotation}. The lower section shows \ours{}, where the model is first fine-tuned with interpretation-aware supervision using misleadingness rationales and then used for rationale-guided correction of misleading previews.}    \label{fig:framework}
    \vspace{-4mm} 
\end{center}
\end{figure*}

\subsection{Data Collection}
\label{sec:data_collection}

To rigorously characterize omission-based misleadingness, a benchmark must meet two criteria: \textbf{(1)} factual reliability, where the full context provides an authoritative ground truth; and \textbf{(2)} narrative complexity, where the story is rich enough for omissions to meaningfully alter interpretation.

To meet these criteria, we build \ds{} on top of VisualNews~\cite{visualnews}, a trustworthy repository of multimodal news reporting, and apply a strict two-stage filtering process to ensure both relevance and difficulty. First, to ensure \textbf{societal relevance}, we stratify sampling across ten high-impact topics (e.g., politics and public health; see Appendix~\ref{app:topic_selection}), covering domains most vulnerable to misinformation. Second, to enforce \textbf{informational density}, we retain instances whose previews can only be properly interpreted with answers to high-level questions such as \textit{``Why did this happen?''} and \textit{``How will it unfold?''} (see Appendix~\ref{app:context_filtering}), and discard simple factual announcements (e.g., dates or locations).

This filtration yields a foundational corpus of high-quality news pairs $(P_{\text{news}}, C_{\text{news}})$, which serve as the input for the multi-stage annotation pipeline detailed in the following subsection.


\subsection{Multi-Stage Data Annotation}
\label{sec:data_annotation}

Motivated by cognitive theories of news comprehension, where readers first form an initial situation model from surface cues before updating it with full text~\cite{entman1993framing}, we leverage the social simulation capabilities of LLMs~\cite{liu2025stepwise} to design a multi-stage annotation pipeline (Figure~\ref{fig:framework}). This pipeline explicitly models the interpretative gap between the preview and the full context.

\textbf{Stage 1: Preview-based Understanding Simulation ($U_p$).}
Simulating a user scrolling through a feed, the LLM is presented strictly with the news preview $P_{news}$ (headline + image). It generates an understanding $U_p$ comprising \textbf{(1)} a surface description of the visible content and \textbf{(2)} a deeper inference of the event’s implications, conditioned solely on these limited cues (see Figure~\ref{prompt:preview_understanding}).

\textbf{Stage 2: Context-based Understanding Simulation ($U_c$).}
To establish the news ground truth, the LLM processes the full article text $C_{news}$ to generate a comprehensive understanding $U_c$. This represents the informed interpretation of a reader who has fully consumed the story (see Figure~\ref{prompt:context_understanding}).

\textbf{Stage 3: Misleading Omission Judgment.}
We detect misleadingness by comparing the semantic divergence between the preview interpretation ($U_p$) and the context interpretation ($U_c$). Specifically, the model evaluates whether the preview omits critical information that alters the perceived event nature, causal relationships, or emotional valence (details in Appendix~\ref{app:misleadinf_judgment}).

\textbf{Quality Control via Cross-Model Agreement.}
To mitigate single-model bias, we use Gemini-2.5-pro and GPT-4.1 as independent annotators and apply strict cross-model agreement filtering, retaining only instances where both models assign the same misleadingness label (see Appendix~\ref{app:cross_model_agreement} for details). This prioritizes label precision over recall, ensuring that our final dataset consists of unambiguous cases. The resulting \ds{} benchmark contains 6,000 instances (5,000 train, 1,000 test), balanced 1:1 between misleading and non-misleading previews.

\subsection{Corrective Headline Generation}
\label{sec:misleading_correction_annotation}

To establish a gold-standard reference for active mitigation, we generate corrected headlines $\hat{T}$ corresponding to the two protocols defined in Section~\ref{sec:misleading_omission_correction}: \textbf{Minimal-Edit} and \textbf{Free-Form}.

To ensure benchmark solvability, we apply a strict \textbf{validity filter}: because not all instances can be corrected successfully, we retain only those corrections that are verified as the ground truth (via the pipeline in Section~\ref{sec:data_annotation}) to successfully eliminate misleadingness under \textit{both} protocols (see details in Appendix~\ref{app:misleading_correction_detailes}). 

Instances that cannot be resolved through headline revision alone, typically because the misleadingness stems from the visual component, are excluded from the primary benchmark and analyzed separately in Section~\ref{exp:multimodal_analysis}.


\subsection{Data Quality Assessment}
\label{sec:data_quality_assessment}
To validate the quality of our automated pipeline, we conduct human evaluation with three independent annotators, assessing detection fidelity on the full test set and correction validity on 100 randomly sampled instances. We evaluate alignment along three dimensions: \textbf{(1) Interpretation Consistency}, which measures whether the simulated ($U_p, U_c$) align with human perception; \textbf{(2) Detection Accuracy}, which evaluates the correctness of the predicted misleadingness labels; and \textbf{(3) Correction Effectiveness}, which assesses whether the revised headlines successfully remove the misleading impression.

The results show strong agreement between our pipeline and expert annotators across all three dimensions, with Fleiss’ $\kappa \ge 0.81$ and accuracy above 94\% (see Appendix~\ref{app:human_eval}). These findings indicate that our automated annotations provide reliable ground truth for both training and evaluation.

\section{\ours{}: Mitigating Misleading Omissions in News Previews}
\label{sec:OMGuard}

To govern omission-based misleadingness, we propose \ours{}, a framework that integrates both detection and mitigation. As shown in Figure~\ref{fig:framework}, \ours{} operates in two stages:
\textbf{(1) Interpretation-Aware Fine-tuning}, which conditions the model on explicit narrative interpretations to strengthen its detection capability; and
\textbf{(2) Rationale-Guided Mitigation}, which uses the model's diagnostic rationales to guide targeted headline correction.

\subsection{Interpretation-Aware Fine-Tuning}
\label{sec:Interpretation-Aware_Fine-Tuning}
We formulate omission-based misleadingness detection in multimodal news previews as an instruction-tuning task. Each training instance is represented as a tuple
$\{I, T, C_{\text{news}}, U_p, U_c, y, r\}$.
The intermediate interpretations ($U_p, U_c$) are provided as \textit{auxiliary reasoning context}, while loss supervision is applied only to the final judgment $y$ and its rationale $r$.

We opt \textit{not} to compute loss on $U_p$ and $U_c$ for two methodological reasons. First, narrative interpretation is inherently non-unique. Enforcing exact token-level alignment risks promoting stylistic imitation rather than helping the model learn a robust decision boundary for misleadingness.
Second, our objective is \textit{logic extraction}. The model should learn how to leverage interpretive discrepancies to justify its conclusion, rather than learn to reproduce a specific set of reference interpretations.

Formally, we maximize the likelihood of the judgment and rationale conditioned on the full interpretive context:
\vspace{-0.7em}
\begin{multline}
\label{eq:fine}
\mathcal{L}_{\text{FT}}
= - \sum_{t=1}^{|s|}\log P_{\theta}(s_t \mid x, s_{<t}), \\
\text{where } s = [r \, ; \, y], 
\ x=\{I, T, C_{news}, U_p, U_c\}.
\end{multline}

\vspace{-0.8em}
\textbf{Inference Pipeline.}
During inference, ground-truth interpretations are not available. We therefore adopt a \textbf{generate-then-reason} process. The model first generates its own preview-based and context-based interpretations ($U_p$ and $U_c$). These generated interpretations are then incorporated into the input to trigger the fine-tuned detection head, which produces the final label $y$ and rationale~$r$.

\subsection{Rationale-Guided Mitigation}
\label{sec:mitigation}

A key advantage of \ours{} is that its detection rationales ($r$) enable \textbf{active mitigation}. Unlike standard black-box classifiers, \ours{} provides a structured diagnosis that identifies the specific omission and clarifies why the preview is misleading relative to the full context.

We use this diagnosis to support zero-shot headline correction. When a preview is identified as misleading, the model receives the generated rationale $r$ together with the correction constraints $\mathcal{C}$ (for example, minimal-edit requirements). Conditioning the rewrite on $r$ directs the model to make targeted adjustments that resolve the omission while preserving the stylistic quality of the original headline, as described in Equation~\ref{eq:mitigation}.



\section{Experiments}
\label{sec:experiments}
\begin{table*}[t]
\renewcommand\arraystretch{1.1}
\setlength{\tabcolsep}{8 pt} 
\small
  \begin{center}
\begin{tabular}{lccccccc}
\hline \hline
  \multirow{2}{*}{\textbf{Models}} & \multirow{2}{*}{\textbf{Accuracy}} &  \multicolumn{3}{c}{\textbf{Non-Misleading}} & \multicolumn{3}{c}{\textbf{Misleading}} \\ \cline{3-8}
  & & \textbf{Precision} & \textbf{Recall} & \textbf{F1} & \textbf{Precision} & \textbf{Recall} & \textbf{F1} \\ \hline
  Llava-7B & 0.50 & 0.50 & 1.00 & 0.67 & 0.00 & 0.00 & 0.00 \\ 
  Qwen3-VL-8B & 0.68 & 0.62 & 0.95 & 0.75 & 0.89 & 0.42 & 0.57 \\ 
  InternVL3.5-8B & 0.64 & 0.59 & 0.95 & 0.73 & 0.86 & 0.34 & 0.49 \\ 
  LLama3-VL-11B & 0.51 & 0.51 & 0.58 & 0.54 & 0.51 & 0.43 & 0.47 \\ \hline
  Qwen3-VL-8B-Thinking & 0.70 & 0.70 & 0.68 & 0.69 & 0.69 & 0.71 & 0.70 \\ 
  GLM-4.1V-9B-Thinking & 0.70 & 0.74 & 0.60 & 0.66 & 0.67 & 0.79 & 0.72 \\ 
  Qwen3-VL-235B-Thinking & 0.79 & 0.77 & 0.81 & 0.79 & 0.80 & 0.76 & 0.78 \\ \hline
  LLama3-VL-90B  & 0.72 & 0.67 & 0.85 & 0.75 & 0.80 & 0.58 & 0.67 \\ 
  Qwen3-VL-235B & 0.86 & 0.80 & 0.96 & 0.88 & 0.96 & 0.76 & 0.85 \\
  \hline
  FT w $U_{p}/U_{c}$ (End-to-End Inference) & 0.87 & 0.83 & 0.94 & 0.88 & 0.93 & 0.80 & 0.86 \\ 
  FT w $U_{p}/U_{c}$ (Multi-stage Inference) & 0.78 & 0.70 & 0.98 & 0.81 & 0.97 & 0.57 & 0.72 \\ 
  FT Label-Only & 0.84 & 0.79 & 0.93 & 0.85 & 0.91 & 0.75 & 0.82 \\ 
  \ours{} (Interpretation-Aware) & 0.86 & \textbf{0.87} & 0.84 & 0.85 & 0.84 & \textbf{0.88} & \textbf{0.86} \\
 \hline \hline
\end{tabular}
\caption{Misleading omission detection effectiveness of LVLMs and \ours{} on \ds{}.}
    \label{tab:misleading_detection}
    \vspace{-4mm}
  \end{center}
\end{table*}

We organize our experiments around the following research questions:

\textbf{RQ1:(\S \ref{exp: misleading_quantify})} To what extent can LVLMs reliably detect omission-induced misleadingness in multimodal news previews?

\textbf{RQ2:(\S \ref{exp:misleading_rewritten})}What factors affect the effectiveness of misleadingness correction in news previews?

\textbf{RQ3:(\S \ref{exp:frame_analysis})} What mechanisms give rise to omission-based misleadingness in news previews?

\textbf{RQ4:(\S \ref{exp:multimodal_analysis})} What visual cues contribute to omission-induced misleadingness?

\subsection{Implementation Details}
\label{exp:implementation_details}


\textbf{Models.} We evaluate 8 representative open-source LVLMs on \ds{}, covering diverse parameter scales and major model families: \textbf{(1) Non-Reasoning models:} Qwen-VL~\cite{qwen3technicalreport} (Qwen3-VL-8B, Qwen3-VL-235B), Llama3-VL~\cite{llama3} (Llama3-VL-11B, Llama3-VL-90B), InternVL3.5-8B~\cite{wang2025internvl3_5}, and LLaVA-7B~\cite{llaVA1.5_7b}; \textbf{(2) Reasoning models:} Qwen-VL~\cite{qwen3technicalreport} (Qwen3-VL-8B-Thinking, Qwen3-VL-235B-Thinking) and GLM-4.1V-9B-Thinking~\cite{hong2025GLM}. We provide detailed model cards in Table.~\ref{tab:model_cards}.

\paragraph{Fine-tuning details.} We fine-tuned Qwen3-VL-8B using LoRA~\cite{hu2022lora}. 
Specifically, we set the LoRA rank to \(r=64\), the scaling factor to \(\alpha=16\), and the dropout rate to 0.05. 
The model was trained for 1 epoch using the AdamW optimizer with a learning rate of \(2 \times 10^{-5}\), and an effective batch size of 4 was achieved via gradient accumulation.

\subsection{Omission Detection Performance}
\label{exp: misleading_quantify}

\textbf{Standard LVLMs under-detect misleading omissions.}
As shown in Table~\ref{tab:misleading_detection}, general open-source models exhibit large disparities in class-wise recall. Although larger models, such as the 90B and 235B variants, achieve higher overall accuracy, they remain biased toward the non-misleading class and often fail to identify subtle omission-based deception. This reveals a persistent blind spot in recognizing the \emph{interpretation drift} between the preview and the full context.

\textbf{Reasoning models raise recall but increase false positives.}
Reasoning-enhanced models improve sensitivity to misleading cues through multi-step processing, but this gain comes at the cost of precision. As shown in Table~\ref{tab:misleading_detection}, these models exhibit \textit{overthinking}~\cite{sui2025stop}: they often amplify minor benign inconsistencies into misleading signals, increasing recall on positive cases while sharply reducing precision. This suggests that generic reasoning mechanisms still require task-specific calibration to avoid excessive false positives.

\textbf{Interpretation-aware fine-tuning improves class balance and robustness.}
We compare \ours{} with a \textit{Label-Only} ablated variant that applies supervision only to the final label $y$. Although both methods achieve accuracy comparable to a 235B model, the \textit{Label-Only} variant yields substantially lower recall on misleading cases. This suggests that label-only supervision encourages reliance on superficial patterns. In contrast, \ours{} uses the interpretive gap as a reasoning anchor, enabling stronger causal discrimination that generalizes better to complex omission scenarios. By explicitly contrasting preview-based and context-based understanding, the model is encouraged to attend to the missing contextual elements that actually drive misleadingness, instead of relying on shallow lexical or visual cues. This leads to a more balanced decision boundary across classes and more robust performance on subtle omission cases. Qualitative examples are shown in Figure~\ref{fig:case_judgment}.

\begin{table*}[t]
\renewcommand\arraystretch{1}
\setlength{\tabcolsep}{12 pt} 
\small
  \begin{center}
\begin{tabular}{lcccccccc}
\hline \hline
  \multirow{2}{*}{\textbf{Rewrite Model}} &  \multicolumn{4}{c}{\textbf{Free-form}} & \multicolumn{4}{c}{\textbf{Minimal-edit}} \\ \cline{2-9} 
  & \textbf{G1} & \textbf{G2} & \textbf{G3} & \textbf{G4} & \textbf{G1} & \textbf{G2} & \textbf{G3} & \textbf{G4}\\ \hline
  Qwen3-VL-8B & 0.88 & 0.84 & 0.84 & 0.27 & 0.84 & 0.78 & 0.78 & 0.25\\
  InternVL-3.5-8B & 0.82 & 0.72 & 0.82 & 0.20 & 0.82 & 0.75 & 0.79 & 0.21\\
  GLM-4.1V-9B-Thinking & 0.82 & 0.76 & 0.84 & 0.59 & 0.73 & 0.66 & 0.73 & 0.51\\ 
  Qwen3-VL-235B & 0.90 & 0.84 & 0.89 & 0.56 & 0.77 & 0.68 & 0.75 & 0.46\\ \hline

  
  OMGuard & \textbf{0.95} & \textbf{0.90} & \textbf{0.95} & \textbf{0.75} & \textbf{0.94} & \textbf{0.90} & \textbf{0.95} & \textbf{0.75} \\
  
 \hline \hline
\end{tabular}
\caption{Correction success rate (CSR) comparison for misleading omission correction.
\textbf{G1:} oracle rationale guided headline correction;
\textbf{G2:} self-generated rationale guided headline correction, applied only to samples the model predicts as misleading;
\textbf{G3:} oracle rationale guided headline correction on the same sample subset as G2;
\textbf{G4:} full end-to-end misleading omission correction pipeline.
Detailed setups are provided in Appendix~\ref{app:experimental_setup}.}
    \label{tab:misleading_correction}
  \end{center}
\end{table*}
\begin{table*}[t]
\renewcommand\arraystretch{1.05}
\setlength{\tabcolsep}{8 pt} 
\small
  \begin{center}
    
\begin{tabular}{lcccccc}
\hline \hline
  \multirow{2}{*}{\textbf{Rewrite Model}} &  \multicolumn{3}{c}{\textbf{Free-Form}} & \multicolumn{3}{c}{\textbf{Minimal-Edit}} \\ \cline{2-7} 
  & \textbf{BLEU-4} & \textbf{ROUGE-L} & \textbf{Cosine} & \textbf{BLEU-4} & \textbf{ROUGE-L} & \textbf{Cosine} \\ \hline

  \textbf{Oracle (GPT-5)} & 7.36 (\texttt{--}) & 23.89 (\texttt{--}) & 0.54 (\texttt{--})
        & 45.51 (\texttt{--}) & 61.54 (\texttt{--}) & 0.77 (\texttt{--}) \\ \hline

  Qwen3-VL-8B &5.43 (4.88) & 22.39 (23.20) & 0.53 (0.52)
              & 9.57 (10.14) & 29.57 (29.89) & 0.59 (0.54) \\
  InternVL3.5-8B & 4.24 (2.65) & 20.98 (18.36) & 0.50 (0.43)
                 & 4.60 (4.16) & 21.30 (20.00) & 0.51 (0.44) \\
  GLM-Thinking & 8.76 (5.97) & 26.54 (25.71) & 0.55 (0.55)
               & 28.24 (24.96) & 47.40 (44.66) & 0.69 (0.65) \\
  Qwen3-VL-235B & 3.27 (4.42) & 24.03 (24.60) & 0.55 (0.56)
               & 27.36 (20.17) & 52.16 (45.30) & 0.75 (0.71) \\
               \hline

  OMGuard & 7.97 (7.50) & 23.73 (22.92) & 0.53 (0.53)
          & 13.15 (11.17) & 28.70 (25.89) & 0.59 (0.55) \\

 \hline \hline
\end{tabular}
\caption{Similarity comparison between corrected headlines $\hat{T}$ and the original misleading headlines $T$ under the \textit{Free-form} and \textit{Minimal-edit} settings. \textbf{Oracle (GPT-5)} denotes the annotated reference correction. Results are reported for two rationale sources: \textbf{oracle rationale} and \textbf{self-generated rationale}.}
    \label{tab:rewritten_quality}
    \vspace{-4mm}
  \end{center}
\end{table*}

\textbf{\ours{} reduces error propagation from imperfect interpretations.}
\begin{figure}[t]
\begin{center}
    \includegraphics[width=1\linewidth]  {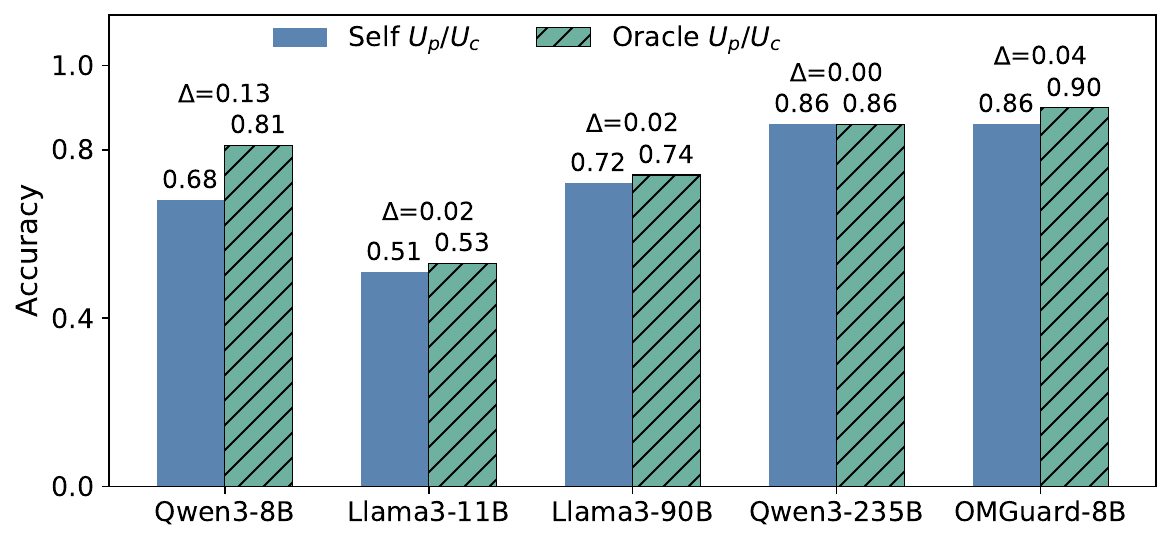}
    \caption{Quantifying error propagation via oracle substitution of $U_p/U_c$; $\Delta=\text{Acc}(\text{oracle }U)-\text{Acc}(\text{self }U)$.}
    \label{fig:U_ablation}
\end{center}
\end{figure}
To assess sensitivity to interpretation quality, we replace the model’s self-generated $U_p$ and $U_c$ with oracle versions produced by GPT-4.1 during annotation and measure the resulting performance gap $\Delta$. As shown in Figure~\ref{fig:U_ablation}, smaller models exhibit larger $\Delta$, indicating stronger error propagation from imperfect intermediate interpretations. By contrast, larger models have $\Delta$ close to zero, suggesting that their main bottleneck lies in recognizing misleadingness itself rather than forming intermediate interpretations. Relative to its backbone, \ours{} reduces $\Delta$ from 0.13 to 0.04, showing that supervising only the final decision $y$ and rationale $r$, without directly supervising $U_p$ and $U_c$, can effectively mitigate error propagation. This supports our design choice to treat $U_p$ and $U_c$ as reasoning context rather than training targets.

\textbf{Direct supervision on $U_p/U_c$ introduces conservative bias.} 
As discussed in Section~\ref{sec:Interpretation-Aware_Fine-Tuning}, \ours{} does not treat $U_p/U_c$ as direct supervision targets during training. To verify this design choice, we directly supervise $U_p/U_c$ and evaluate both end-to-end and multi-stage inference. As shown in Table~\ref{tab:misleading_detection}, the end-to-end setting biases the model toward the non-misleading class, increasing precision to 0.93 but reducing misleading recall to 0.80. This suggests that the model adopts an overly strict threshold for predicting misleadingness and becomes less sensitive to subtle omission-based cases unless its generated reasoning is sufficiently explicit. Under multi-stage inference, direct supervision on intermediate steps further entangles content understanding with misleadingness discrimination, producing a more severe train--test mismatch. As a result, overall accuracy drops to 0.78, and misleading recall declines sharply to 0.57.

\textbf{Separating explanation from judgment improves deployment suitability.}
Although end-to-end inference is simpler and performs slightly better on some precision-oriented metrics, separating explanation from final judgment remains the better design for deployment. A multi-stage framework provides greater controllability, intervention, and auditability. End-to-end inference makes it difficult to inspect or correct flawed intermediate understanding before the final decision is made, whereas a multi-stage pipeline explicitly exposes intermediate interpretations and supports targeted intervention, such as adding corrective context or human feedback. This decoupled design also better supports integration of auxiliary signals, such as user comments, verified background information, and community notes, which is especially valuable for real-world content moderation and governance.

\subsection{Omission Mitigation Performance}
\label{exp:misleading_rewritten}

We evaluate the active mitigation capability of \ours{} using \textbf{Correction Success Rate} (CSR) across diverse editing modes and setups, and measure headline quality against ground-truth corrections with BLEU, ROUGE, and cosine similarity; see Appendix~\ref{app:experimental_setup} for details.

\textbf{Correction of misleading previews reveals a substantial perception--correction gap.}
As shown in Table~\ref{tab:misleading_correction}, a substantial gap emerges when comparing G1 and G4. With oracle rationales (G1), even open-source LVLMs achieve high rewriting success rates (average 0.86 for Free-form and 0.72 for Minimal-edit), demonstrating strong inherent correction capability. In contrast, the full end-to-end pipeline (G4), which requires detection, rationale generation, and correction, reaches much lower success rates (average 0.41 for Free-form and 0.36 for Minimal-edit). This indicates that the primary bottleneck lies in the upstream detection stage. Although fine-tuning improves detection and reduces the gap between G1 and G4, the gap does not fully close, suggesting that rationale quality and other downstream factors also influence final correction performance.

\textbf{High-quality rationales are essential for effective correction.}
Comparing G2 and G3 in Table~\ref{tab:misleading_correction} shows that, among correctly detected misleading instances, higher-quality judgment rationales consistently yield higher CSR (avg. Free-Form: 0.79 to 0.85; Minimal-Edit: 0.72 to 0.76). Consistently, Table~\ref{tab:rewritten_quality} indicates that oracle rationales more often produce rewrites that are semantically faithful to the source and better aligned with the ground truth than those using self-generated rationales. We further remove rationales and provide only the \textit{binary misleading label}; CSR drops to 17\%, confirming that labels alone do not provide sufficient guidance. Overall, effective correction critically depends on accurately localizing the misleading content and explaining why it is misleading.

\textbf{Editing constraints expose a trade-off between control and success rate.}
As shown in Table~\ref{tab:rewritten_quality}, the stronger constraints in the Minimal-Edit setting make instruction following substantially more difficult. Reasoning-enhanced and large-parameter models remain closer to the oracle references, but their limited room for modification often lowers CSR. Other models achieve higher apparent success only by violating the editing constraints, as reflected in their lower BLEU and ROUGE scores under Minimal-Edit. Under Free-Form editing, relaxing these constraints reduces CSR differences across models. Even so, models with stronger instruction-following ability remain closer to the oracle references, producing more controlled rewrites with smaller deviations. Overall, \ours{} achieves a better balance between adherence to editing constraints and correction effectiveness, resulting in more stable performance across both settings.

\textbf{\ours{} transfers from diagnosis to correction without rewrite supervision.}
Although \ours{} is fine-tuned only for detection and rationale generation, and receives no direct supervision on rewriting, it still delivers substantial gains in end-to-end correction. This supports our central hypothesis: improving the model’s ability to \textit{diagnose} subtle interpretive discrepancies through Interpretation-Aware Fine-Tuning naturally strengthens its ability to \textit{correct} them. As a result, \ours{} enables effective zero-shot mitigation without explicit training on headline rewriting.

\subsection{Attributions of Misleading Omissions}
\label{exp:frame_analysis}

We analyze how previews induce misleadingness from two complementary perspectives: \textbf{frame shift} and \textbf{fine-grained attribution}. First, following~\cite{arora2025multi_framing}, we extract the top-3 frames from both the preview and the full context and compute their overlap ratio (see Appendix~\ref{app:frame_analysis}). Second, we identify major categories of misleadingness through expert annotation: two human experts first summarize five recurring causes, and an LLM judge then assigns an attribution label to each sample (see Appendix~\ref{app:Fine-grained misleadingness attribution}).

\begin{figure}[t]
\begin{center}
    \includegraphics[width=\linewidth]  {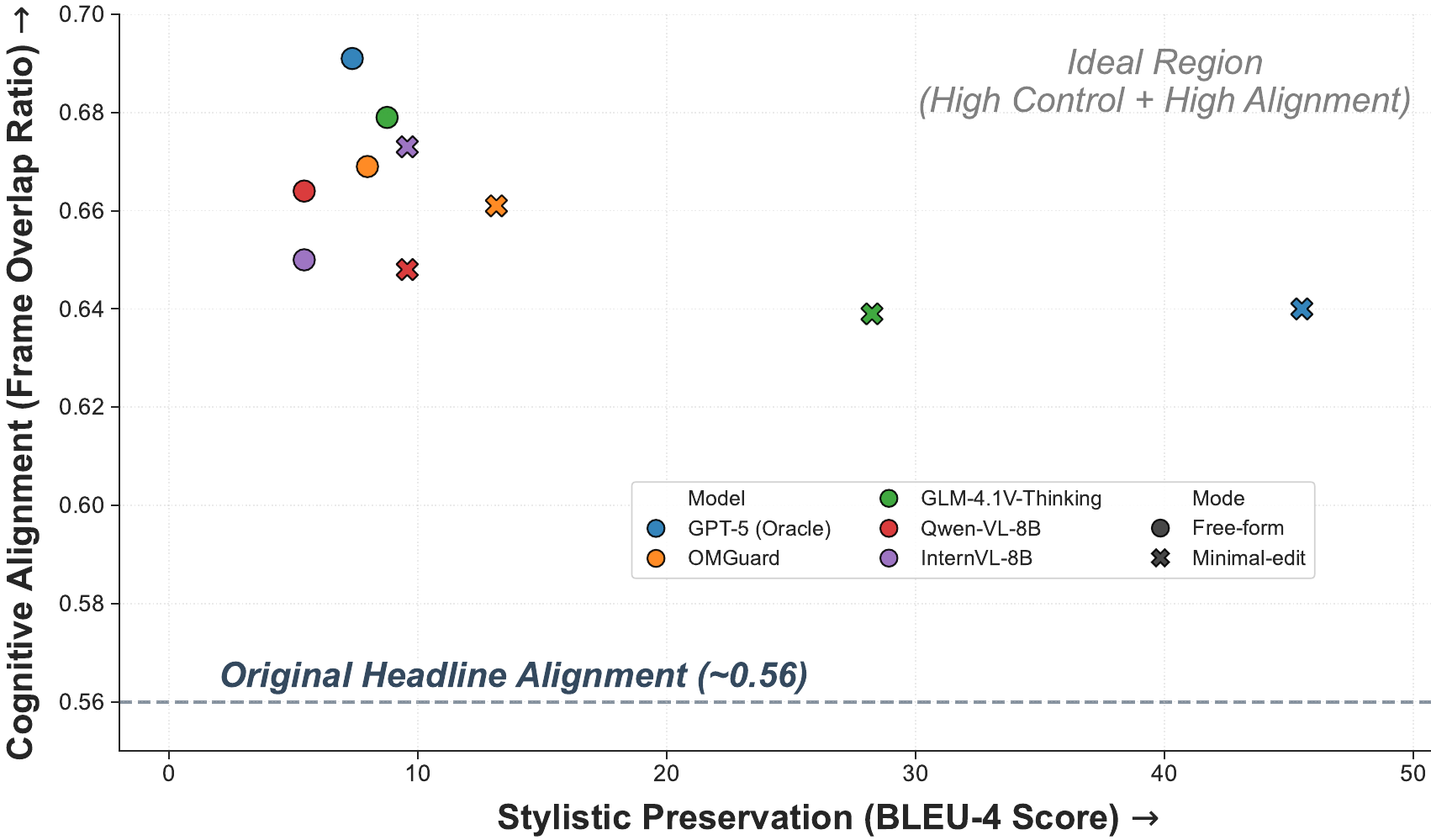}
    \caption{Frame-shift analysis comparing stylistic preservation and frame alignment in misleading headline correction.}
    \label{fig:frame_analysis}
\end{center}
\end{figure}

\textbf{Misleading omissions usually arise from local narrative shifts.}
As shown in Figure \ref{fig:frame_analysis}, the frame overlap ratio between misleading headlines and full contexts remains relatively high ($\approx 0.56$), suggesting that omission-based misleadingness rarely changes the topic itself, but instead distorts its perspective or implication. Although correction improves frame alignment, it does not fully close the gap, indicating that \textit{truthfulness} and \textit{framing} are related but distinct dimensions. As shown in Figure~\ref{fig:frame_analysis}, \textit{Minimal-Edit} better preserves stylistic continuity but shifts the frame less effectively, whereas \textit{Free-Form} achieves stronger alignment at the cost of the original stylistic voice.

\textbf{Missing background is the dominant source of misleadingness.} We further examine fine-grained causes beyond frame shift. For misleading samples in the test set, we first ask human experts to summarize five major causes, and then use an LLM-as-a-judge approach~\cite{llm-as-a-judge, mllm-as-a-judge} to assign an attribution label to each sample (see Appendix~\ref{app:Fine-grained misleadingness attribution}). 

\begin{figure}[t]
\begin{center}
    \includegraphics[width=\linewidth]  {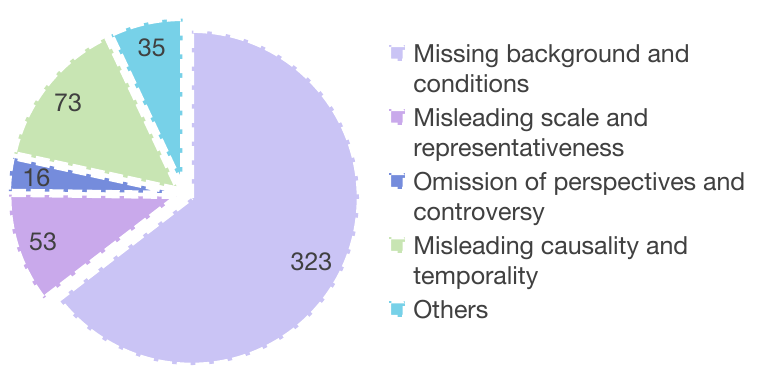}
\caption{Distribution of misleadingness types.}
    \label{fig:fine_misleading_attribution}
\end{center}
\end{figure}

The resulting distribution follows a long-tail pattern. As shown in Figure~\ref{fig:fine_misleading_attribution}, \textit{missing background and conditions} accounts for 67\% of all misleading cases, far exceeding the other categories. This distribution also helps explain why effective correction often depends on \emph{recovering omitted context} instead of extensive rewriting. For the dominant omission-driven cases, minimally adding the missing background or conditions is often sufficient to prevent incorrect inferences. By contrast, long-tail categories such as \textbf{misleading causality/temporality} and \textbf{misleading scale/representativeness} often require the rewrite to make implicit relations or scope conditions explicit, and, when possible, to ground the revision more clearly in supporting evidence. Overall, these findings reinforce our central claim that omission-based misleadingness is driven primarily by \emph{insufficient information and selective presentation}: seemingly factual previews can still induce biased interpretations by withholding critical conditions, even without introducing explicit falsehoods.

\begin{table}[t]
\renewcommand\arraystretch{1.1}
\setlength{\tabcolsep}{3 pt} 
\small
  \begin{center}
\begin{tabular}{lcccc}
\hline \hline
 Data type & Precision & Recall & F1 & Accuracy \\ \hline
 Non-Misleading & 0.84 & 0.91 & 0.87 & \multirow{2}{*}{0.87} \\
 Misleading & 0.90 & 0.82 & 0.86 & \\ \hline \hline
\end{tabular}
\caption{Ablation study using only the headline (Text-Only) for misleadingness detection, showing the contribution of visual signals.}
    \label{tab:image_ablation}
  \end{center}
\end{table}

\subsection{Discussion: Multimodal Mitigation}
\label{exp:multimodal_analysis}

\textbf{Visual cues are essential for robust misleadingness detection.}
To assess the need for multimodal modeling, we ablate the image channel and evaluate a headline-only variant while keeping the rest of the pipeline unchanged. As shown in Table~\ref{tab:image_ablation}, removing visual information degrades overall performance and introduces a clear asymmetry across classes: the model retains relatively high recall on non-misleading instances but struggles more on misleading ones. This suggests that many failures arise from \textit{text--image dissonance}, and that reliable detection requires multimodal evidence.


\textbf{Visual semantics limit the effectiveness of text-only correction.}
As noted in Section~\ref{sec:misleading_correction_annotation}, not all misleading instances can be resolved during ground-truth construction. To better characterize this bottleneck, we partition misleading samples into \emph{Text-Fixable} and \emph{Image-Driven} subsets via modality attribution (see Appendix~\ref{app:image_correction_analysis}). Even before attribution, the Free-Form rewriting success rate reaches only 0.78 (Appendix~\ref{app:misleading_correction_detailes}); after attribution, performance rises to 0.88 on the text-fixable subset but falls to 0.46 on image-driven cases. This sharp gap highlights the limiting role of visual semantics in omission correction.


\vspace{0.3em}
\textbf{Image intervention can mitigate otherwise uncorrectable cases.}
To test whether \emph{image-driven misleading instances} can be resolved through visual intervention, we conduct a pilot study on 20 cases that text modification alone cannot correct. We use a generative model to synthesize alternative images better aligned with the full context (Figure~\ref{fig:image_modify}), increasing the correction success rate to 90\%. Although we do not advocate replacing authentic news photos with generated images in practice, this result highlights \textit{image selection} as a critical variable in governing misleading previews. More practically, it suggests that future newsroom tools should help editors select more representative archival images, instead of relying solely on headline edits.


\vspace{0.2em}
\section{Conclusion}
\vspace{0.2em}

We present a formal study of \textbf{misleading omissions} in multimodal news previews and introduce \ds{}, a 6,000-instance benchmark for evaluating both detection and correction. We further propose \ours{}, which uses interpretation-aware fine-tuning to lift an 8B model to the detection accuracy of 235B-scale LVLMs while enabling zero-shot headline correction through diagnostic rationales. Our analyses show that most misleadingness arises from local narrative shifts that can often be resolved through targeted text editing, while a smaller set of image-driven cases exposes the limits of text-only correction and motivates multimodal intervention. These findings position omission-based misleadingness as a distinct governance challenge beyond fact verification alone. We hope \ds{} and \ours{} provide a foundation for future work on multimodal moderation, newsroom support, and more context-faithful communication online.
\clearpage

\section*{Limitations}
Our work also has the following limitations: (1) We primarily mitigate preview-level misleadingness by rewriting news headlines. While this text-based correction can effectively recover critical missing information and reduce erroneous inferences, rewriting itself may introduce new biases, such as shifts in emphasis or unintended formulations driven by the model’s inherent biases. Ensuring mitigation without introducing new distortions is an important and promising direction, but it is beyond the scope of this paper. (2) We conduct image-side visual prototyping. However, our current exploration is largely limited to generating semantically aligned images to validate feasibility, and we do not investigate more practical solutions, such as retrieving candidate images from newsroom archives or related media repositories. Retrieval-based real-image replacement and joint text–image adjustment frameworks remain promising avenues for future work.

\section*{Ethical Considerations}
Our work aims to model and identify reader-understanding deviations induced by a news preview relative to the full news context, and to mitigate misleading impressions caused by information omission and selective presentation through headline rewriting. Overall, this direction has the potential to reduce erroneous inferences in low-information preview settings and to lessen the societal harms associated with the spread of misleading content, thereby supporting content governance and healthier information consumption.

We also recognize potential misuse risks. First, the detection and correction capabilities themselves could be exploited adversarially. Attackers may treat misleadingness detection as an optimization signal to iteratively craft omission-based misleading content that is harder to detect, or to perform evasion-oriented optimization against our evaluation setup. Second, our exploration of visual prototyping is intended to highlight the importance of image selection in governance, but similar techniques could be misused to generate or curate more suggestive images that amplify misleading impressions. Our goal is to promote responsible governance practices, not to provide tools for generating deceptive content.

\section*{Acknowledgments}
This work is supported by the Yunnan Research Project (Grant Nos. 202503AG380006, 202401AT070474, 202501AU070059, and 202403AP140021), the National Natural Science Foundation of China (Grant Nos. 62562061, 62502422, and 62462067), the Yunnan Provincial Department of Education Science Research Project (Grant Nos. 2025J0006, 2024J0010, and 2025J0007) and China Scholarship Council (CSC) program. This research is also supported by the Ministry of Education, Singapore, under its Academic Research Fund Tier 1 (T1 251RES2508) and MOE AcRF TIER 3 Grant (MOE-MOET32022-0001).

\bibliography{custom}

\appendix


\appendix

\clearpage

\appendix

\section{Appendix}
\label{sec:appendix}

\subsection{Related Work}
\label{app:related_work}
\paragraph{Evidence-based Fact Checking.} Capitalizing on the powerful capabilities of large language models (LLMs) across various tasks~\cite{wang2026mmcomet, ma2026attention, compselect, zhang2026stable}, recent studies~\cite{qi2024sniffer, li-etal-2025-cmie, he2025factguard, wu2025beyond,li2025imrrf,li2025drifting, zeng2026manipulation, ma2025graphing} predominantly rely on LLMs augmented with retrieved external evidence to evaluate whether image–text pairs align with real-world facts. SNIFFER~\cite{qi2024sniffer} fine-tunes LVLMs and incorporates external cues to verify authenticity. CMIE~\cite{li-etal-2025-cmie}, E2LVLM~\cite{wu2025e2lvlm}, and DiFaR~\cite{wan2025difar} improve cross-modal alignment and optimize evidence utilization. LRQ-FACT~\cite{lrq-fact} and LEMMA~\cite{xuan2024lemma} focus on enhancing retrieval to obtain more targeted evidence, thereby improving verification performance.Recent studies have also examined omissions, but primarily in the form of contextual omission. TRACER~\cite{tracer} and OmiGraph~\cite{OmiGraph} retrieve missing information to infer implied intent, yet they largely treat such omissions as auxiliary cues for fact-checking and are mostly confined to plain-text settings.

Rather than verifying the factual veracity of the news preview, we focus on scenarios in which it is factually correct yet omits key information relative to the full news context, thereby inducing a shift in interpretation.

\paragraph{News Preview Understanding Beyond Veracity.}
Recent studies have also examined how the \emph{presentation} of news itself shapes audience understanding and dissemination outcomes. DeceptionDecoded~\cite{wu2025seeing} adopts a generative perspective to analyze how news creators construct misleading presentations under different intents, while NINT~\cite{NINT} and InSide~\cite{Inside} leverage intent as an auxiliary signal to strengthen misinformation detection. Meanwhile, journalism research commonly regard frames~\cite{de2005news} as a concrete manifestation of selective emphasis and omission, and existing work~\cite{arora2025multi_framing,moernaut2020visual,lucking2012framing,ali2022survey} primarily focuses on identifying and attributing frame categories. In addition, studies on clickbait and curiosity-driven headlines~\cite{wang2025multi, yu2024multimodalclick, hagen2022clickbait}, analyze the effects of news presentation by examining how exaggerated, suspenseful, and information-incomplete previews attract clicks and influence click-through rates and user engagement.

Different from our work, these studies typically do not characterize the interpretation shift induced by a preview relative to its underlying full context. In contrast, we formalize misleadingness as a reader-side interpretation shift and further introduce a correction mechanism to mitigate the potential misleadingness caused by news previews.

\subsection{Data Annotation Details}
\subsubsection{Topic Selection}
\label{app:topic_selection}
VisualNews~\cite{visualnews} provides fine-grained topic annotations for each news article. Leveraging these annotations, we stratify sampling across ten high-impact topic categories to improve \textbf{societal relevance} and ensure the benchmark emphasizes domains most susceptible to misinformation.

The selected topics are:

\begin{tcolorbox}
[colback=black!2!white,colframe=white!50!black,boxrule=0.5mm]
"world", \\
"international\_relations",  \\
"politics\_elections", \\
"politics",  \\
"law\_crime", \\
"business\_economy",  \\
"environment", \\
"science\_technology", \\
"technology",  \\
"conflict\_attack"
\end{tcolorbox}

\subsubsection{Content Filtering}
\label{app:context_filtering}
In Section \ref{sec:data_collection}, we discussed to enforce \textbf{information density} via content filtering. We filter out low-quality instances (e.g., cases where the headline is simply a direct caption of the image) and focus the datas on semantically more complex news reports. We retain instances where the full story raises and answers complex high-level inquiries, such as \textit{``what happened'', ``why it happened'', and ``how will it unfold''}. The detailed filtering strategy is outlined in Figure~\ref{prompt:content_filtering}.

\subsubsection{Misleading Omission Judgment}
\label{app:misleadinf_judgment}
Section~\ref{sec:data_annotation} details our procedure for determining whether a news preview $P_{\text{news}}$ is misleading based on $U_p$ and $U_c$. Concretely, we compare the initial impression formed from the image--headline pair alone ($U_p$), including the event’s nature and status, causal relations, responsibility attribution, and perceived severity, with the judgment after reading the full context ($U_c$). We label a preview as \textit{misleading} when the full context substantially revises or overturns these core judgments, rather than merely refining or supplementing them. Otherwise, when the context mainly adds details while preserving the overall direction of the core judgment, we label it as \textit{Non-misleading} (Figure~\ref{prompt:misleading_judgment}).

\subsubsection{Details of Cross-Model Agreement}
\label{app:cross_model_agreement}
In Section~\ref{sec:data_annotation}, we describe our cross-model agreement filtering strategy to mitigate single-model bias and improve annotation reliability. Based on our analysis, rationales $r$ produced by the GPT family better match human writing style, so we first use GPT-4.1~\cite{openai2025gpt4_1} for multi-stage annotation. We then have Gemini-2.5-pro~\cite{google2025gemini_2_5} annotate the same instances independently and apply agreement filtering. The two models produce consistent labels for over 84\% of the samples. Because our goal is to identify omission-based misleading content, we retain only high-confidence instances where both models agree, prioritizing label precision over coverage of borderline cases.

\subsubsection{Details of Misleadingness Correction}
\label{app:misleading_correction_detailes}
In Section~\ref{sec:misleading_correction_annotation}, we describe how we construct a gold-standard reference for actively mitigating misleading instances. Specifically, for each misleading sample, we use the stronger GPT-5 to generate two corrected versions under the two settings of $\mathcal{C}$, namely \textsc{Minimal-Edit} and \textsc{Free-Form}, and Figure~\ref{prompt:headline_correction} illustrates the detailed procedure. We then follow the annotation protocol in Section~\ref{sec:data_annotation} to re-assess whether $\hat{P}_{news}=(\hat{T}, I)$ remains misleading. Finally, to obtain a high-confidence reference set that balances correctability and structural controllability, we retain only instances that are successfully corrected under both settings. In the test split, which contains 500 misleading instances, 78\% can be corrected under \textsc{Free-Form}, whereas only 54\% can be corrected under \textsc{Minimal-Edit}. Ultimately, we select the 265 instances that are successfully corrected under both $\mathcal{C}$ settings as the gold-standard reference.

\subsubsection{Human Evaluation Details}
\label{app:human_eval}
\begin{figure*}[t!]
\begin{center}
    \includegraphics[width=\linewidth]  {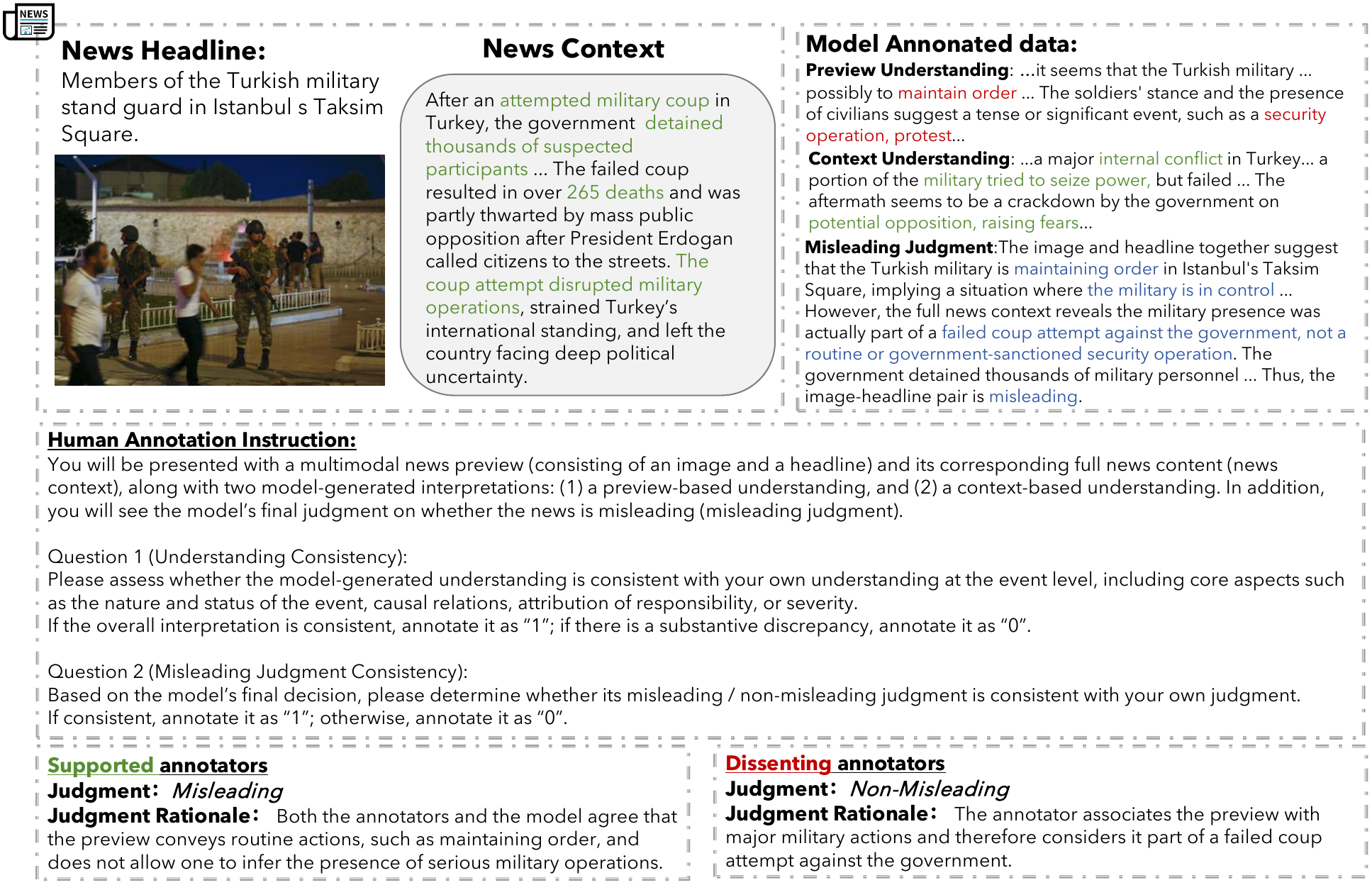}
    \caption{Human annotation guidelines for misleading content detection and an analysis of annotation disagreements.}
    \label{fig:detection_inconsistency}
\end{center}
\end{figure*}

\begin{figure*}[t!]
\begin{center}
    \includegraphics[width=\linewidth]  {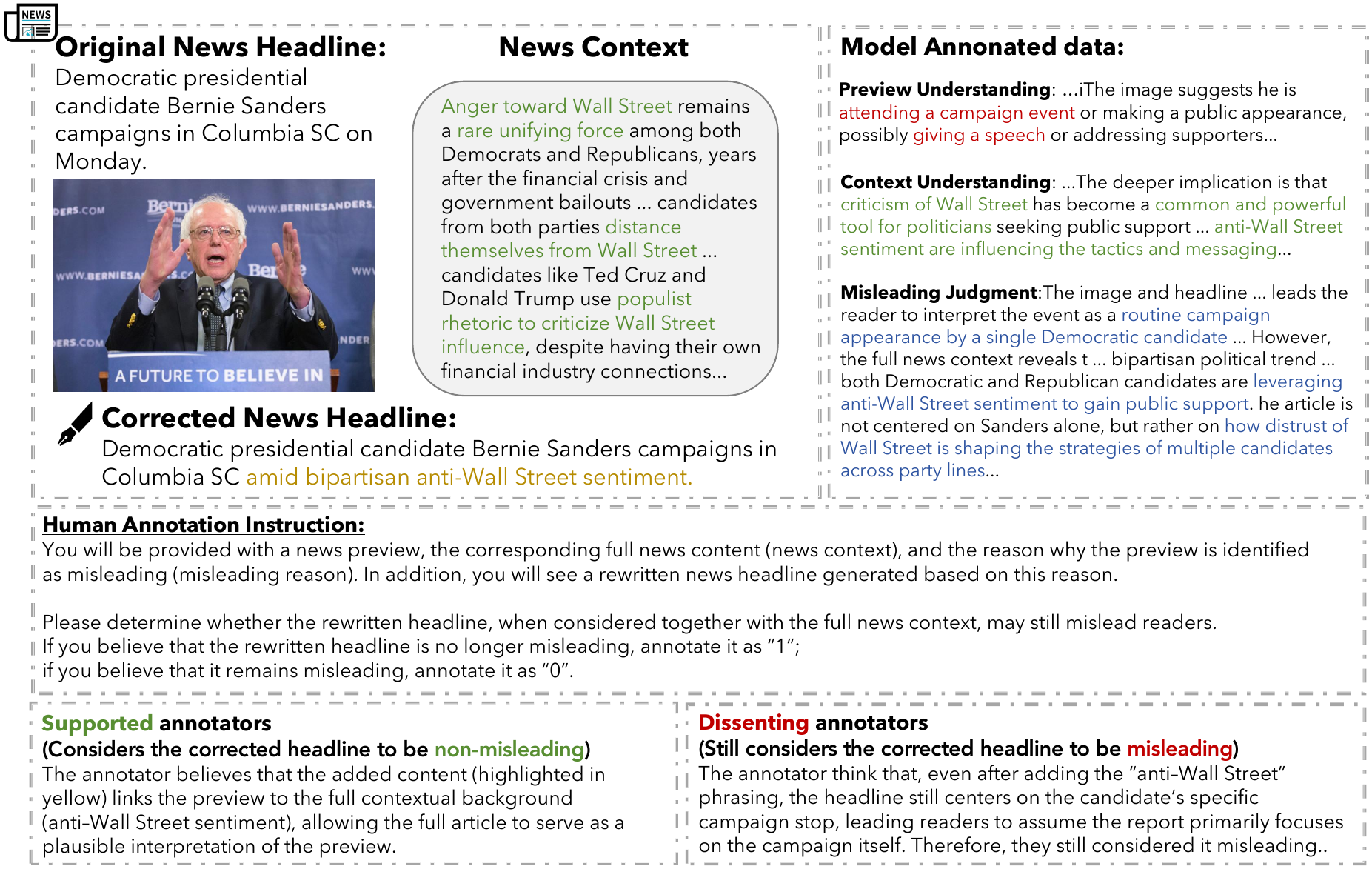}
    \caption{Human annotation guidelines for misleading content correction and an analysis of annotation disagreements.}
    \label{fig:rewritten_inconsistency}
\end{center}
\end{figure*}

In Section~\ref{sec:data_quality_assessment}, we describe our human evaluation. We assess annotation quality using accuracy and inter-annotator agreement. For the misleadingness detection task, we evaluate both (i) understanding consistency and (ii) judgment consistency of the model outputs. Accuracy is defined as whether the model-simulated understanding or the final misleadingness judgment matches the human annotators’ decisions, and the evaluation protocol is shown in Figure~\ref{fig:detection_inconsistency}. For the headline correction task, accuracy is defined as whether human annotators judge the rewritten headline to be no longer misleading, and the evaluation protocol is shown in Figure~\ref{fig:rewritten_inconsistency}.

As shown in Table~\ref{tab:human_eval}, we report detailed validation results. Here, \textit{$(U_p, U_c)$} corresponds to the evaluation of understanding consistency, \textit{Detection} corresponds to judgment consistency, and \textit{Correction} corresponds to the headline correction task. The results indicate that our automated annotations achieve high agreement with expert judgments.

\begin{table}[h]
\renewcommand\arraystretch{1.2}
\setlength{\tabcolsep}{5 pt} 
\small
  \begin{center}
\begin{tabular}{lcccc}
\hline 
 Metrics & $U_p$ & $U_c$ & Detection & Correction \\ \hline
 Accuracy & 0.95 & 0.96 & 0.94 & 0.95 \\
 Fleiss'~$\kappa$ & 0.83 & 0.86 & 0.81 & 0.82 \\ \hline 
\end{tabular}
\caption{Automated annotations exhibit high agreement with expert judgments across interpretation consistency, detection accuracy, and correction effectiveness.}
    \label{tab:human_eval}
  \end{center}
\end{table}

In addition, we analyze cases with annotator disagreement. For misleading omission detection, disagreements primarily stem from annotators adopting different event frames or inferred implications when interpreting the same image--headline preview, which in turn yields different judgments after reading the full context, as illustrated in Figure~\ref{fig:detection_inconsistency}. For headline correction, while rewrites typically add or amend key information, some annotators may not view the added content as salient, or may apply different thresholds for whether the rewrite sufficiently eliminates misleadingness, leading to inconsistent judgments, as shown in Figure~\ref{fig:rewritten_inconsistency}.

\subsection{Experimental Details}
\label{app:exp_details}

\subsubsection{Experimental Setup}
\label{app:experimental_setup}
We provide detailed definitions of G1--G4 in Table~\ref{tab:misleading_correction} as follows:
\begin{itemize}
\item \textbf{G1 (Oracle Upper Bound):} For all samples with gold-standard references, we prompt open-source LVLMs to rewrite the headline, using the \emph{Rationale} produced during annotation as guidance for correction. This setting estimates whether the open-source LVLM can perform effective correction when provided with oracle rationales.
\item \textbf{G2 (Self-Rationale Guidance):} For all samples with gold-standard references, we first require open-source LVLMs to detect misleading omissions. For the subset predicted as \textit{misleading}, we further evaluate the model’s ability to perform headline correction using its self-generated rationale.
\item \textbf{G3 (Oracle-Rationale Guidance):} Using the same samples as G2, we replace the guidance with the gold-standard \emph{Rationale} and perform headline rewriting, thereby quantifying the effect of rationale quality on correction performance.
\item \textbf{G4 (End-to-End):} The full pipeline from detection to rationale generation to rewriting. For all samples with gold-standard references, the LVLM first predicts whether the preview is \textit{misleading}. If it is deemed \textit{misleading}, the model then performs headline correction conditioned on its self-generated \emph{rationale}. This setting reflects real-world governance performance.
\end{itemize}

For misleading content correction, we evaluate performance using the correction success rate, defined as the proportion of misleading instances that are relabeled as non-misleading after correction. Given the corrected headlines, we re-assess misleadingness using the same evaluation model and multi-stage pipeline as in data annotation, ensuring a fair comparison of success rates across different rewrite models. In addition, we report BLEU, ROUGE, and cosine similarity to quantify the gap between constraint-based rewrites and the oracle headline.

\subsection{Misleading Attribution Analysis}
\subsubsection{Frame Shift Analysis}
\label{app:frame_analysis}
In Section~\ref{exp:frame_analysis}, we examine frame shifts between misleading and non-misleading previews. We adopt the set of generic news frames from~\cite{arora2025multi_framing}, which we list below, and Figure~\ref{prompt:frame_detection} outlines the frame identification procedure. Specifically, we extract three semantic frames from the original preview, the rewritten preview, and the corresponding full news context. We then quantify frame shift by counting the number of overlapping frames between each preview (before and after rewriting) and the full context.
The generic news frames:

\begin{tcolorbox}
[colback=black!2!white,colframe=white!50!black,boxrule=0.5mm]
``Economic'', \\
``Capacity and Resources'', \\
``Morality'', \\
``Fairness and Equality'', \\
``Legality'', \\
``Policy'', \\
``Crime and Punishment'', \\
``Security and Defense'', \\
``Health and Safety'', \\
``Quality of Life'', \\
``Cultural Identity'', \\
``Public Opinion'', \\
``Political'', \\
``External Regulation'', \\
``Other''
\end{tcolorbox}

\subsubsection{Fine-grained Misleadingness Attribution}
\label{app:Fine-grained misleadingness attribution}
In Section~\ref{exp:frame_analysis}, we further analyze fine-grained factors beyond frame shift that can induce misleadingness. For misleading samples in the test set, we first ask human experts to summarize five major causes, and then adopt an LLM-as-a-judge approach~\cite{llm-as-a-judge, mllm-as-a-judge} to assign an attribution label to each sample. Figure~\ref{prompt:fine_grained_misleading_attribution} provides definitions and explanations of these causes, along with the full attribution pipeline. As shown in Figure~\ref{fig:fine_misleading_attribution}, \textbf{Missing background and conditions} dominates the distribution (about 65\%, 323/500), while the remaining causes exhibit a clear long-tail pattern. This suggests that most misleading previews do not negate the core facts. Instead, they exploit \emph{missing qualifiers}, such as temporal scope, applicability constraints, comparison baselines, subsequent developments, entity identities, or causal premises. When such prerequisites are omitted, readers are prone to extrapolate a local description into a global conclusion, leading to systematic bias.

\subsection{Multimodal Mitigation}


\subsubsection{Image-Modal Analysis for Correction}
\label{app:image_correction_analysis}

\paragraph{Modality Attribution for Correction Failure:} 
In Section~\ref{exp:multimodal_analysis}, to examine whether correction performance is significantly affected by the image modality, we first identify a news preview as misleading and then, based on its misleading rationale, use an LLM judge to categorize it as either \emph{Text-Fixable} (misleadingness can be corrected by rewriting the headline) or \emph{Image-Driven} (misleadingness is primarily driven by the image and cannot be corrected by headline rewriting alone). We then compare correction outcomes across these two subsets to quantify the impact of the image modality. The detailed prompt are provided in Figure \ref{prompt:modality_attribution} and definitions of \textit{Text-Fixable} and \textit{Image-Driven} are as follows:

\begin{tcolorbox}
[colback=black!2!white,colframe=white!50!black,boxrule=0.5mm]
Text-Fixable: \\
- If the misleading effect mainly stems from information omission, missing outcome, or omitted controversy in the headline, and the image itself merely serves as scene or atmosphere rendering—without anchoring a narrative, identity, event type, or timeline that is fundamentally inconsistent with the main theme of the article—then the case is considered “headline amendable.” In such cases, the misleading impression can be eliminated by rewriting the headline. \\ 

Image-Driven: \\
- If the image content strongly dominates the reader’s interpretation, anchoring an event type, emotion, identity, causality, or historical timeline that is seriously inconsistent with the true news context—even when the headline is maximally revised—the misleading effect cannot be corrected. Such cases are considered “not amendable,” and require image replacement or other multimodal interventions.
\end{tcolorbox}

\paragraph{Visual Prototyping:} 

In Section~\ref{exp:multimodal_analysis}, we conduct a pilot study on 20 “uncorrectable” instances and perform visual prototyping by using a generative model to synthesize alternative images that are more semantically consistent with the full context. Concretely, we know the reason why each preview is misleading, and we also have the failed corrected headline along with the reason it remains misleading. This allows us to diagnose why headline-only edits are insufficient. More importantly, it enables us to infer what image semantics would mitigate the misleading impression when paired with the preview, and to derive prompts that would produce such images. We then use GPT to generate the corresponding images from these prompts. The full generation pipeline is shown in Figure~0, and the generated images together with their mitigation effects are presented in Figure~\ref{fig:image_modify}.

We note that the generated images in this experiment may not be fully factual, and our misleadingness assessment does not additionally verify their factual correctness. Therefore, our goal is not to produce factually accurate images, but to validate a key point: pairing news previews with semantically appropriate images can substantially reduce potential misleadingness. In practice, before publishing a news preview, creators can use similar semantic guidance to retrieve and select semantically matched real images from a repository. Accordingly, newsroom tools should assist editors in choosing more representative archival images to reduce misleading risk, rather than relying solely on headline edits.

\begin{figure*}[t!]
\begin{center}
    \includegraphics[width=\linewidth]  {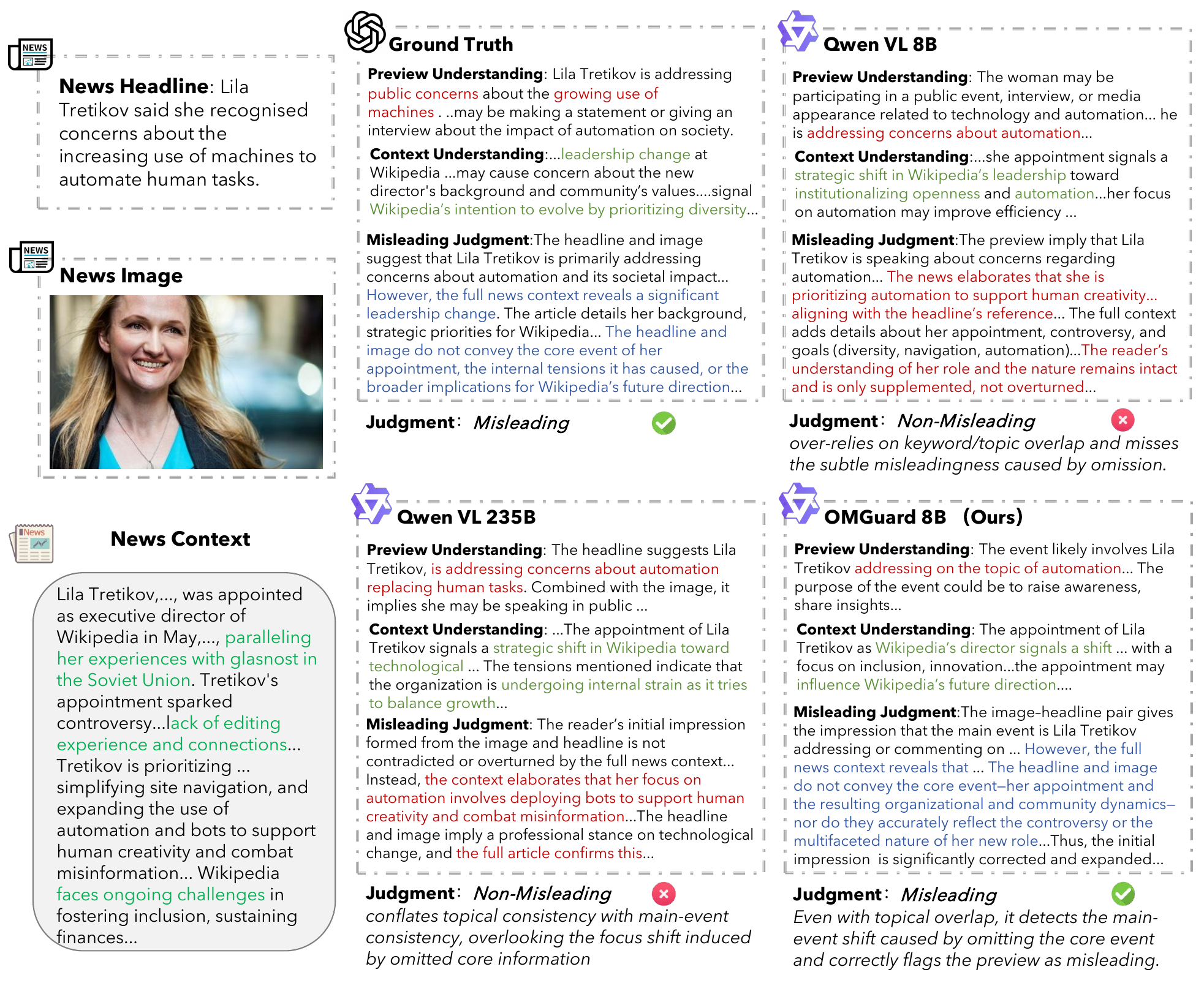}
    \caption{A case study of misleading content detection across different models.}
    \label{fig:case_judgment}
\end{center}
\end{figure*}

\begin{figure*}[t!]
\begin{center}
    \includegraphics[width=\linewidth]  {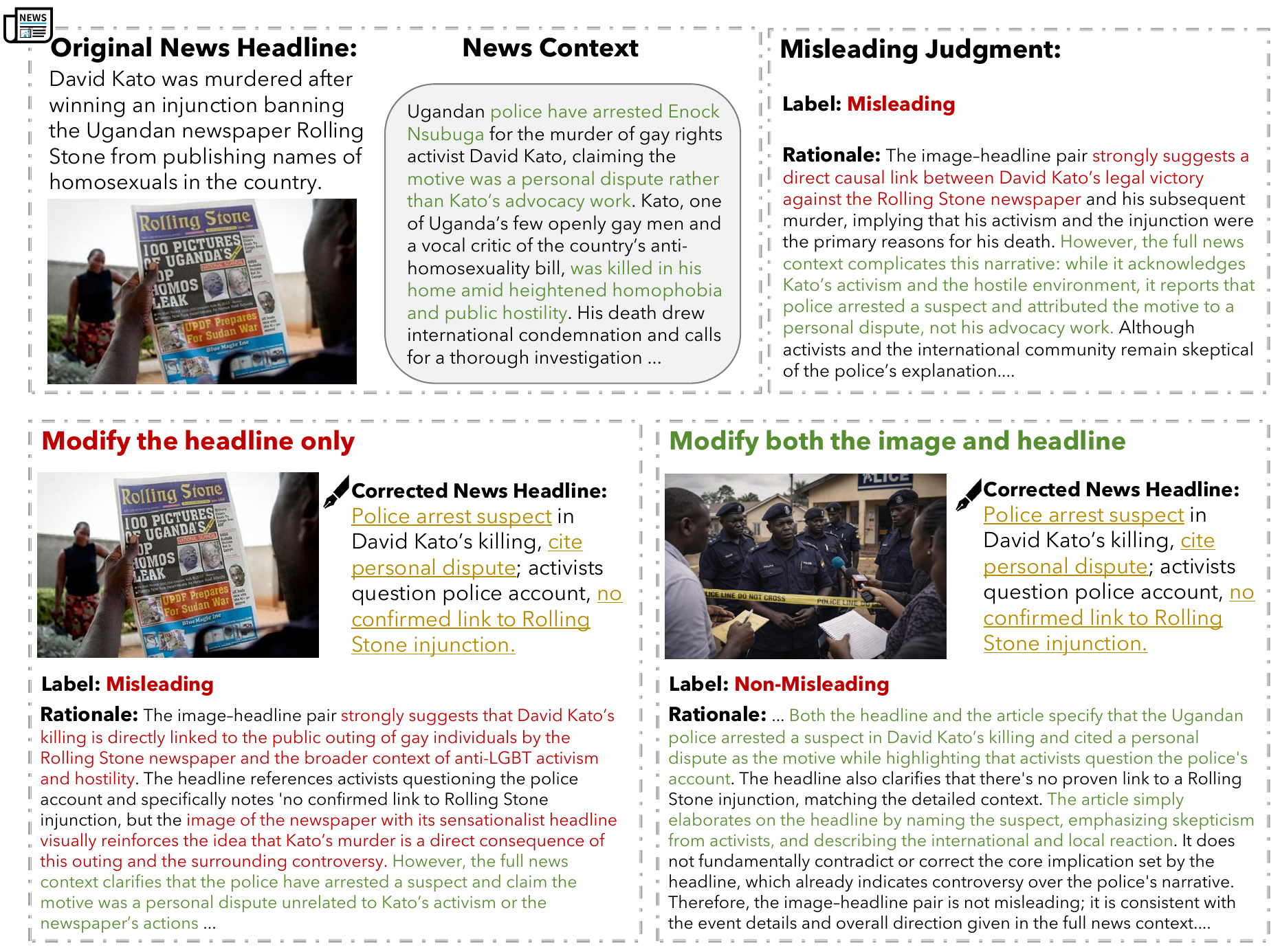}
    \caption{Examples of correction via visual image replacement. We show that replacing the original image with a generated suggestion image that better matches the article context can effectively mitigate misleadingness in news previews.}
    \label{fig:image_modify}
\end{center}
\end{figure*}

\begin{figure*}
\begin{prompt}{Context Filtering}
\small  

You task is to annotate image–text pairs for news-signal screening. \\

Definitions: \\
- ld (Literal-Descriptive): venue/object/scene or general information (e.g., opening, decor, menu, discount). Does not convey any event information worth further inquiry; readers would not ask ``what happened/why/what’s next.'' \\
- ms (Message-Suggestive): conveys or implies a real-world event/impact (e.g., conflict, explosion, bombing, disaster, accident, evacuation, casualties, law enforcement, arrest, protest, policy, curfew, sanctions), or uses causal/temporal language that invites ``what happened/why/what’s next.'' \\

\# Input \\
IMAGE: You will be provided with the image. \\
TEXT: \{News Headline\} \\

\# Output Requirements
Outputs must follow the JSON format below, consisting of three keys: \\

``label'': ``ld''| ``ms'', \\
``reason'': ``rationale citing the main textual cues''

\end{prompt}
\caption{Prompt for selecting high-quality news instances.}
\label{prompt:content_filtering}
\end{figure*}
\begin{figure*}
\begin{prompt}{LLM-based Preview Understanding Simulation}
\small  
\# Task Description \\

You are an average news reader. you will be provided with a piece of news that includes an image and a news headline.  
From a reader’s perspective, describe your immediate impression of the news and make reasonable inferences at the detail level. \\

You need to complete the following parts: 

   - Analyze **only** based on the image and the news headline.   \\
   - Describe what you see (surface interpretation).  \\
   - Infer what event might be happening based on visual cues and the headline (event implication).   \\ 
  
\# Input 

News Headline: \{NEWS\_HEADLINE\} 

Image: (will be provided) 
 \\ 
 
\# Output Format (JSON)

    ``Image–Headline'': \{ \\ 
        ``Surface\_Interpretation'': ``What is the surface interpretation?'', \\ 
        ``Event\_Implication'': ``What is the deep meaning, and what is the purpose?'' 
    \}

\end{prompt}
\caption{Prompt for LLM-based preview understanding simulation.}
\label{prompt:preview_understanding}
\end{figure*}
\begin{figure*}
\begin{prompt}{LLM-based News Context Understanding Simulation}
\small  
\# Task Description 

You are an average news reader. you will be provided with a full news article.  
From a reader’s perspective, describe your immediate impression of the news and make reasonable inferences at the detail level. \\

You need to complete the following parts: 

   - Analyze based on the news context.   \\
   - Describe what you see (surface interpretation).  \\
   - Infer what event might be happening based on the news context (event implication).   \\ 
  
\# Input \\ 
News Context: \{NEWS\_CONTEXT\} \\ 

\# Output Format (JSON) 
 
 ``News\_Context'': \{ \\ 
        ``Surface\_Interpretation'': ``What is the surface interpretation?'', \\ 
        ``Event\_Implication'': ``What is the deep meaning, and what is the purpose?'' 
    \}

\end{prompt}
\caption{Prompt for LLM-based news context understanding simulation.}
\label{prompt:context_understanding}
\end{figure*}
\begin{figure*}
\begin{prompt}{Misleading Omission Judgment}
\small  
\# Task Description \\

You will receive an image, a news headline, a full news context, a reader’s surface interpretation and event implication for the image–headline pair, a reader’s surface interpretation and event implication for the full news context. \\

You need to complete the following parts: 

- If a reader forms an impression about the nature, status, cause and effect, the responsible party, or severity of a news event when only exposed to images and titles, and this impression is significantly corrected, restricted, or overturned after reading the full news, it is considered misleading.

- On the contrary, if the full news only elaborates, extends, or supplements the content implied by the title (for example, by providing more details, reactions, or outcomes), without altering the reader's understanding of the basic direction or core judgment of the event, it is considered non-misleading. \\
  
\# Input 

Image: (will be provided) 

News Headline: \{NEWS\_HEADLINE\} 

Full News Context: \{CONTEXT\} 

Reader Interpretations based on image-headline and context: \{READER\_INFER\} \\ 

\# Output Format (JSON) \\
 
\{
    ``Misleading'': ``Yes/No'',
    ``Reason'': "Not less than 100 words, focus on the event level.''
    \}

\end{prompt}
\caption{Prompt for misleading omission judgment.}
\label{prompt:misleading_judgment}
\end{figure*}
\begin{figure*}
\begin{prompt}{News Headline Correction}
\small  

\# Task Description \\

You are a news rewriting expert. You will receive an news image, an news headline, and the full news context. Compared with the news context, the image-headline pair is considered misleading. You will also be provided with the corresponding reason why it is misleading. \\

Please follow the steps below to generate a non-misleading headline:

1. Analyze the Misleading Cause 
   - Based on the provided data, identify the main reasons why the original headline is misleading, including any factual, contextual, or expressive distortions.

2. Suggestions on Improvement 
   - Consider what kinds of information or phrasing should be included in the headline to prevent misleading readers and accurately convey the core message of the news.

3. Generate the Headline
   - Based on the above analysis, produce a non-misleading headline that is factually accurate, semantically clear, and maintains a neutral tone. \\

\# Rewriting requirements:

\textbf{[This can be replaced according to different rewritten types]}

\textbf{Minimal-Edit:}
- The rewritten news headline may contain at most \{limit\_words\} additional words compared to the original headline.
- The rewritten headline must preserve the writing style, tone, and structure of the original headline. \\

\textbf{Free-Form:}
- The rewritten news headline may contain at most \{limit\_words\} additional words compared to the original headline. \\

\# Input:
Image: You will be provided.

News Headline: \{NEWS\_HEADLINE\}

Full News Context: \{NEWS\_CONTEXT\}

Misleading reason of image-headline pair: \{MISLEADING\_REASON\} \\

\# Output(json): \\
\{ 
``Misleading\_Cause'': xxx,
``Suggested\_Improvement'': xxx,
    ``Rewritten\_Caption'': xxx
\}

\end{prompt}
\caption{Prompt for misleading headline correction.}
\label{prompt:headline_correction}
\end{figure*}
\begin{figure*}
\begin{prompt}{News Frame Identification}
\small  

\# Task Definition\\
You are an expert media analyst. Your task is to identify the relevant generic news frames presented by the combination of the provided News Image and Headline. \\

\# Instruction: \\
- Analyze how the image and headline interact. A news item often contains multiple angles (e.g., both "Political" and "Policy"). 

- Select the **Top-3 most relevant frames** that represent the dominant perspectives from the taxonomy: \{taxonomy\} \\

\# Input \\
IMAGE: You will be provided with the image. \\
TEXT: \{News Headline\} \\

\# Output  \\
Output strictly in JSON format with two keys: \\
\{
- ``reasoning'': Brief explanation of why these frames apply. \\
- ``frames'': A list of strings containing the exact names of the selected frames (e.g., ["Economic", "Political", "Policy"]). \}

\end{prompt}
\caption{Prompt for frame analysis.}
\label{prompt:frame_detection}
\end{figure*}
\begin{table*}[t]
\centering
\small
\begin{tabular}{l|c}
\toprule
\textbf{Model} & \textbf{Model Card} \\
\midrule
GPT-4.1~\cite{openai2025gpt4_1}  & \texttt{gpt-4.1-2025-04-14}\\
GPT-5~\cite{openai2025gpt5}  & \texttt{gpt-5-2025-08-07} \\
Gemini-2.5-Pro~\cite{google2025gemini_2_5}  & \texttt{gemini-2.5-pro-preview-03-25} \\
\midrule
LlaVA-1.5-7B~\cite{llaVA1.5_7b} & \texttt{llava-1.5-7b-hf} \\
Qwen3-VL-8B~\cite{qwen3technicalreport} & \texttt{Qwen3-VL-8B-Instruct} \\
InternVL3.5-8B~\cite{wang2025internvl3_5} & \texttt{InternVL3\_5-8B} \\
Llama3-VL-11B~\cite{llama3} & \texttt{Llama-3.2-11B-Vision-Instruct} \\
GLM-4.1V-9B-Thinking~\cite{hong2025GLM} & \texttt{GLM-4.1V-9B-Thinking } \\
Qwen3-VL-8B-Thinking~\cite{qwen3technicalreport}  & \texttt{Qwen3-VL-8B-Thinking } \\
Llama3-VL-90B~\cite{llama3} & \texttt{Llama-3.2-90B-Vision-Instruct} \\
Qwen3-VL-235B~\cite{qwen3technicalreport} & \texttt{qwen3-vl-235b-a22b-instruct} \\

\bottomrule
\end{tabular}
\caption{Model cards for the LVLMs used in our work.}
\label{tab:model_cards}
\end{table*}
\begin{figure*}
\begin{prompt}{Fine-grained Misleading Attribution}
\small  
\# Task
You are a misleading attribution classifier, designed to evaluate the reasons why an image–headline pair may be misleading compared to the full news context.
Your task is to determine which category of misleading type the given reason belongs to. \\

\# Input \\
- Image: You will be provided. \\
- News Headline: \{NEWS\_HEADLINE\} \\ 
- Full NEWS Context: \{NEWS\_CONTEXT\} \\ 
- Reason why an image–headline pair may be misleading compared to the full news context: \{REASON\} \\

\# Categories \\ 
Choose exactly one of the following categories: \\

1. Missing background and conditions: \\  
   - The reason mainly points out that the image–headline pair omits essential background or conditions needed to correctly understand the event (for example, prior context, policy constraints, key actors, follow-up developments, or outcomes). Because this context is missing, readers are likely to form an incomplete or distorted overall impression.

2. Misleading scale and representativeness: \\  
   - The reason mainly emphasizes that the image–headline pair misleads about how large, frequent, or systemic the event is. It only shows isolated or local cases, or uses extreme examples in a way that underplays or exaggerates the true scale, prevalence, or impact described in the full news context.

3. Omission of perspectives and controversy: \\  
   - The reason mainly highlights that the image–headline pair hides important viewpoints or controversy. It presents only one side (for example, an official or dominant narrative) while omitting affected groups, opposition voices, counter-arguments, or social conflict that are present in the full news context, leading to a one-sided understanding.

4. Misleading causality and temporality: \\  
   - The reason mainly concerns incorrect or misleading suggestions about cause–effect relations, event sequence, or current status. The image–headline pair may imply that one action directly caused an outcome, that an event is still ongoing, or that a past event is current, in ways that are not supported by the full news context.

5. Others: \\  
   - Use this category if the reason does not clearly fall into any of the four types above, or if you are not confident which category is most appropriate. \\

\# Output

Return the output in standard JSON format with the following fields:

\{ 
  ``attribution\_class'': ``Only the most possible class'', \\
  ``attribution\_reason'': ``Explain in detail why it belongs to this category, referring to the given text for analysis'' \}

\end{prompt}
\caption{Prompt for fine-grained misleading attribution.}
\label{prompt:fine_grained_misleading_attribution}
\end{figure*}
\begin{figure*}
\begin{prompt}{Modality Attribution}
\small  
You will be provided with an image, the corresponding headline, the full news article, a reader interpretation based solely on the image–headline pair, a reader interpretation based on the full news article, and an explanation of why the image–headline pair is considered misleading compared to the complete news content. \\

\# Task 

In multimodal news data, there exist a large number of samples in which the image–headline pair does not align with the main theme of the article context, easily misleading readers. In practice, simply rewriting the headline (text) does not always eliminate this misleading effect. In some cases, the image strongly dominates the narrative focus, emotion, or scene, so even after the headline is maximally revised, readers may still form an understanding that does not match the true news context. Therefore, the goal of this task is to automatically identify and annotate which misleading samples are likely to become non-misleading solely through headline rewriting. \\

\# Judgment Criteria

Text-Fixable: \\
- If the misleading effect mainly stems from information omission, missing outcome, or omitted controversy in the headline, and the image itself merely serves as scene or atmosphere rendering—without anchoring a narrative, identity, event type, or timeline that is fundamentally inconsistent with the main theme of the article—then the case is considered “headline amendable.” In such cases, the misleading impression can be eliminated by rewriting the headline. \\
Image-Driven: \\
- If the image content strongly dominates the reader’s interpretation, anchoring an event type, emotion, identity, causality, or historical timeline that is seriously inconsistent with the true news context—even when the headline is maximally revised—the misleading effect cannot be corrected. Such cases are considered “not amendable,” and require image replacement or other multimodal interventions. \\

\# Input: \\
Image: You will be provided. \\
News Headline: \{NEWS\_HEADLINE\} \\
Full News Context: \{NEWS\_CONTEXT\} \\
reader interpretation based only on the image–headline pair: \{Reader\_Preview\}

a reader interpretation based on the full news article, and an explanation of why the image–headline pair is considered misleading compared to the complete news context: {{Reader\_context}}

Misleading reason of image-headline pair: \{MISLEADING\_REASON\}

\# Output(json): \\
\{
   "label": Text-Fixable or Image-Driven,
   "reason": xxx \}

\end{prompt}
\caption{Prompt for modality attribution.}
\label{prompt:modality_attribution}
\end{figure*}
\begin{figure*}
\begin{prompt}{Visual Prototyping}
\small  

You will receive a news preview (including an image and a headline) and the corresponding news context. It is known that this news preview is misleading compared to the news context. \\

We have rewritten the headline based on the identified original misleading rationale. However, the rewritten headline is still misleading. We believe this is mainly because the image introduces misleading cues. \\

I will provide you with:

Image: (will be provided)

Headline: \{HEADLINE\}

Context: \{CONTEXT\}

Original Misleading Rationale: \{Original Misleading Rationale\}

Rewritten Headline: \{Rewritten Headline\}

Rewritten Misleading Rationale: \{Rewritten Misleading Rationale\} \\

You need to perform visual prototyping: analyze what kind of contextual image the rewritten headline should be integrated with so that the new preview (New Image + Rewritten Headline) is no longer misleading. You should output a description of the recommended image and an image prompt for generating it. \\
Output (JSON): \\
\{
  ``Image description'': ``xxx'',
  ``Image Prompt'': ``xxx''\}

\end{prompt}
\caption{Prompt for visual prototyping.}
\label{prompt:visual_prototyping}
\end{figure*}

\end{document}